\documentclass[runningheads]{llncs}
\usepackage{graphicx}
\usepackage{array}
\usepackage{longtable}
\usepackage{color}
\usepackage{amsmath}

\usepackage[utf8]{inputenc}
\usepackage{graphicx}
\usepackage{pgfplots}
\pgfplotsset{compat=1.14}

\newcommand{\be}{\begin{eqnarray}}
\newcommand{\ee}{\end{eqnarray}}
\newcommand{\bee}{\begin{eqnarray*}}
\newcommand{\eee}{\end{eqnarray*}}

\newcommand{\matrixb}{\left[ \begin{array}}
\newcommand{\matrixe}{\end{array} \right]}

\begin{document}
%


\title{Network Intrusion Detection based on LSTM and Feature Embedding}
%
%
\author{Hyeokmin Gwon, Chungjun Lee, Rakun Keum, and Heeyoul Choi}
\authorrunning{Gwon et al.}
%
\institute{School of Computer Science and Electrical Engineering \\
Handong Global University, Pohang, South Korea 37554\\
\email{\{hank.gwon, ok800726, fkrdns2, heeyoul\}@gmail.com}
}
%
\maketitle              
\begin{abstract}
Growing number of network devices and services have led to increasing demand for protective measures as hackers launch attacks to paralyze or steal information from victim systems. Intrusion Detection System (IDS) is one of the essential elements of network perimeter security which detects the attacks by inspecting network traffic packets or operating system logs. While existing works demonstrated effectiveness of various machine learning techniques, only few of them utilized the time-series information of network traffic data. Also, categorical information has not been included in neural network based approaches. In this paper, we propose network intrusion detection models based on sequential information using long short-term memory (LSTM) network and categorical information using the embedding technique. We have experimented the models with UNSW-NB15, which is a comprehensive network traffic dataset. The experiment results confirm that the proposed method improve the performance, observing binary classification accuracy of 99.72\%.

\end{abstract}

\section{Introduction}
Growing number of network devices and services have increased the importance of network security. There is an increasing demand for protective measures as hackers launch attacks to paralyze or steal information from computer systems connected to network. Examples of these attacks are transmission of malicious files and exploitation of security vulnerability of targets \cite{Beaugnon2018}. In such attacks, hackers interact with the target system, generating network activity \cite{Shah2018}. Network Intrusion Detection System (NIDS) is one of the essential elements of network perimeter security which analyzes these activities and raises alarms \cite{Northcutt2005}. Specifically, NIDS analyzes the header and payload data of incoming and outgoing network packets, and it invokes alerts when detecting a malicious network activity \cite{Laskov2008}.

There are two approaches for detection of malicious network activities: traditional methods and machine learning based ones \cite{Garcia2009}. Both of them involve feature extraction stage, but differ in how to identify malicious activity. In the feature extraction approach, individual packets in network activities are summarized into high level events such as sessions \cite{Zeek}. Each summarized record consists of feature values that characterize the high level event \cite{Lee1999}. Then, in traditional development of NIDS, security experts identify patterns of attacks, deciding threshold ranges for each features \cite{Garcia2009}. On the other hand, in the machine learning based approach, a given model automatically learns patterns of malicious activities from a given dataset \cite{Beaugnon2018}. Recently, machine learning based methods have been attracting more attention over traditional methods, due to its potential capability to detect more complicated patterns in a large scale dataset \cite{Tang2018}.  

While many researchers have experimented various machine learning techniques, time-series information of network traffic data have not received much attention \cite{Beaugnon2018,Tang2018}. As network activity occurs in timely manner, usage of sequential information in machine learning models should lead to more comprehensive analysis as long as the model has enough computational capacity for such additional information. Recurrent neural networks (RNNs) can capture temporal dependence in data, which brought significant advances in the fields of speech recognition and machine translation \cite{Moon2015,Bahdanau2015}, and long short-term memory (LSTM) \cite{Hochreiter1997} or gated recurrent unit (GRU) \cite{Cho2014} are popular RNNs \cite{Greff2015}.

In addition to temporal dependence, categorical information has been neglected in neural network based NIDS. Categorical information means non-numeric (or symbolic) features like protocol type, state, and service in network traffic data. While such features are crucial in recognizing malicious pattern activity, traditional neural network approaches could not accept them as input. Categorical features are very common in natural language processing (NLP), because words are symbols, and there are several feature embedding (or word embedding) techniques \cite{Pennington2014,hchoi2017csl} to handle symbolic words in NLP tasks, like language model and neural machine translation \cite{Mikolov2011,Bahdanau2015}. 

In this paper, we propose to apply LSTM and feature embedding to build intrusion detection models, where sequential information of network traffic data is captured by LSTM and categorical features are utilized with feature embedding. For evaluation, after checking open datasets for network intrusion detection, we use the UNSW-NB15 dataset \cite{Moustafa2015}, which is an up-to-date dataset for network intrusion detection. We assume that records are arranged in timely order, which is to capture temporal dependence for intrusion detection. In experiments, we present that LSTM can effectively model sequential structures for NIDS and feature embedding can make categorical features available in the neural network models. Finally, after many experiments with various options and hyper parameters, LSTM with feature embedding leads to significant improvement in detection performance compared to other machine learning techniques. We have achieved binary classification accuracy of 99.72\% over the UNSW-NB15 dataset. 

The rest of this paper is organized as follows. Related works and backgrounds are described in Section 2. In Section 3, we propose new network intrusion detection models. The experiment results are presented and analyzed in Section 4, followed by Section 5 where we conclude the paper.

\section{Background}

\subsection{Network Intrusion Detection Data}
As an open dataset, we use the UNSW-NB15 dataset, which is a broad-gauge network intrusion detection dataset. UNSW-NB15 was created for standardized evaluation of NIDS \cite{Moustafa2015}. Especially, it aimed to replace the KDD Cup 99 and NSL-KDD datasets, which have been popular datasets for NIDS over the years, but do not convey newly emerging network attack behaviors. As specified in \cite{Moustafa2015}, in order to reflect contemporary hacking behaviors, attacks in UNSW-NB15 were generated using IXIA Perfect Storm, which can simulate attacks listed in CVE website. After arranging a testbed environment with the attack generator, traffic was captured by TCP dump. Then the final dataset was formulated by conducting feature extraction with tools such as Bro and Argus.

\subsection{Network Intrusion Detection Method}

IDS has two detection mechanisms according to definitions of malicious activity \cite{Garcia2009}. Signature-based detection mechanism defines malicious activities, and recognizes behaviors that match the attacks. In contrast, anomaly-based detection mechanism defines normal activity, and recognizes behaviors that deviate from the normal ones. The former mechanism is more compatible with attacks that are already known and shows low false-positive rate compared to the latter. On the other hand, the latter has potential to recognize unknown attacks, but it can suffer from high false-positive rate.

About the two detection mechanisms, there are corresponding development approaches for NIDS: expert-centered and machine learning based ones \cite{Garcia2009}. In expert-centered approach, signatures are written by security experts. For example, Snort, a renowned open-source project for NIDS, lets user write rules by which it examines network packets and creates alerts \cite{Northcutt2005}. This approach requires expert knowledge or rule sets. In the latter approach, on the other hand, signatures are automatically learned by machine learning models. Also, it requires a dataset which contains massive amount of data and corresponding labels specifying attack type of each datum \cite{Beaugnon2018}. 

As pointed out in \cite{Niyaz2015,Beaugnon2018}, some challenges should still be addressed for deployment of machine learning based NIDS to real network environments. In addition, experiment over an open dataset assumes that network traffic can be pre-processed in advance \cite{Lee1999}. Nevertheless, experiments are useful for evaluation of potential detection performance of different machine learning techniques.

After publication of UNSW-NB15, there have been many research works to apply myriad of machine learning techniques to the dataset. Suleiman et al. applied various classical machine learning algorithms such as Random Forest, K-nearest neighbor, and Support Vector Machine \cite{Suleiman2018}. Among the experiments, J48 and K-NN algorithms were proposed as the most suitable models with high efficiency and accuracy. Moustafa et al. experimented anomaly-based detection method based on geometric area analysis using trapezoidal area estimation \cite{Moustafa2017}. Meanwhile, Papamarztivanos et al. proposed a novel approach to NIDS with genetic algorithm and decision tree \cite{Papamartzivanos2018}. In their work, they used genetic algorithm to produce detection rules that compose a decision-tree model. The resulting model was experimented over UNSW-NB15, and it showed good performance in detecting both attacks that are common and rare in the dataset.

Recently, Tama et al. experimented effectiveness of deep neural networks (DNNs) for NIDS on UNSW-NB15 \cite{Tama2017}. In addition, VinayaKumar et al. carried out comparative analysis of DNN models and classical machine learning algorithms \cite{Vinayakumar2019}. After conducting extensive parameter search for optimization, they concluded that DNNs were suitable for development of IDS. Furthermore, Nawir et al. discovered Average One Dependence Encoder (AODE) achieves high accuracy with relatively short amount of classification time \cite{Nawir2018}. 

While previous works demonstrated various machine-learning-based NIDS, most of them did not pay attention to sequential information. As exceptions, Staudemeyer applied long short-term memory (LSTM) network to the KDD99 dataset to improve classification performance \cite{Staudmeyer2015}. In addition, Kim et al. experimented an LSTM-RNN model on KDD99 and obtained significant performance improvement \cite{Kim2016}. However, their experiments were performed based on the KDD99 dataset, which does not reflect contemporary attack behaviors. Moreover, the previous LSTM models used categorical features as if the features are continuous ordinal data. Categorical features need feature embedding like word embedding in NLP \cite{Guo2016}. 

\subsection{Long Short-Term Memory}
In this section, we briefly review recurrent neural network (RNN) and LSTM. For more information, the readers are referred to  \cite{Hochreiter1997,HChoi2018neuro,Staudmeyer2015,Olah2015}.

\begin{figure}[h]
\centerline{\hbox{ 
\includegraphics[width=2.4in]{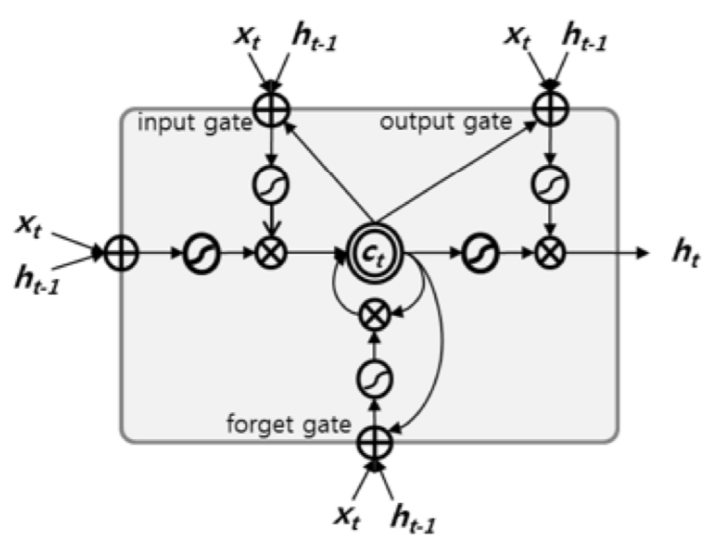}\hspace{0.3in}
}}
\caption{Structure of LSTM Cell. Adopted from \cite{Moon2015}}
\label{fig:LSTM}
\end{figure}

RNN is a modified version of neural network that updates its internal state over time. By forming circular connections within the network, RNNs can memorize past inputs and capture temporal properties in sequential data. However, RNNs can retain memory for only a short amount of time steps due to the vanishing gradient problem. LSTM \cite{Hochreiter1997} solves the vanishing gradient problem by introducing three gates (input, forget and output) around special memory units called cell states $c_t$. The gates control the update of the cell states as shown in Fig. \ref{fig:LSTM}. 

In LSTM, inferences for the cell states $c_t$ and hidden states $h_t$ are given by
\begin{eqnarray}
c_t &=& f_t \odot c_{t-1} + i_t \odot tanh(Uh_{t-1} + Wx_t + b), \\
h_t &=& o_t \odot tanh(c_t),
\end{eqnarray}
where $\odot$ indicates the element-wise multiplication operation, and the three gates are defined by
\begin{eqnarray}
i_t &=& \sigma(W_i x_t + U_i h_{t-1} + b_i), \\
f_t &=& \sigma(W_f x_t + U_f h_{t-1} + b_f), \\
o_t &=& \sigma(W_o x_t + U_o h_{t-1} + b_o).
\end{eqnarray}
Here, $\sigma$ is the sigmoid function, and $U$, $W$, and $b$ are parameters. For implementation convenience without degradation of performance as recommended in \cite{Greff2015}, peephole connections are not included.

\subsection{Word Embedding}
\begin{figure}[h]
\centerline{\hbox{ 
\includegraphics[width=3.0in]{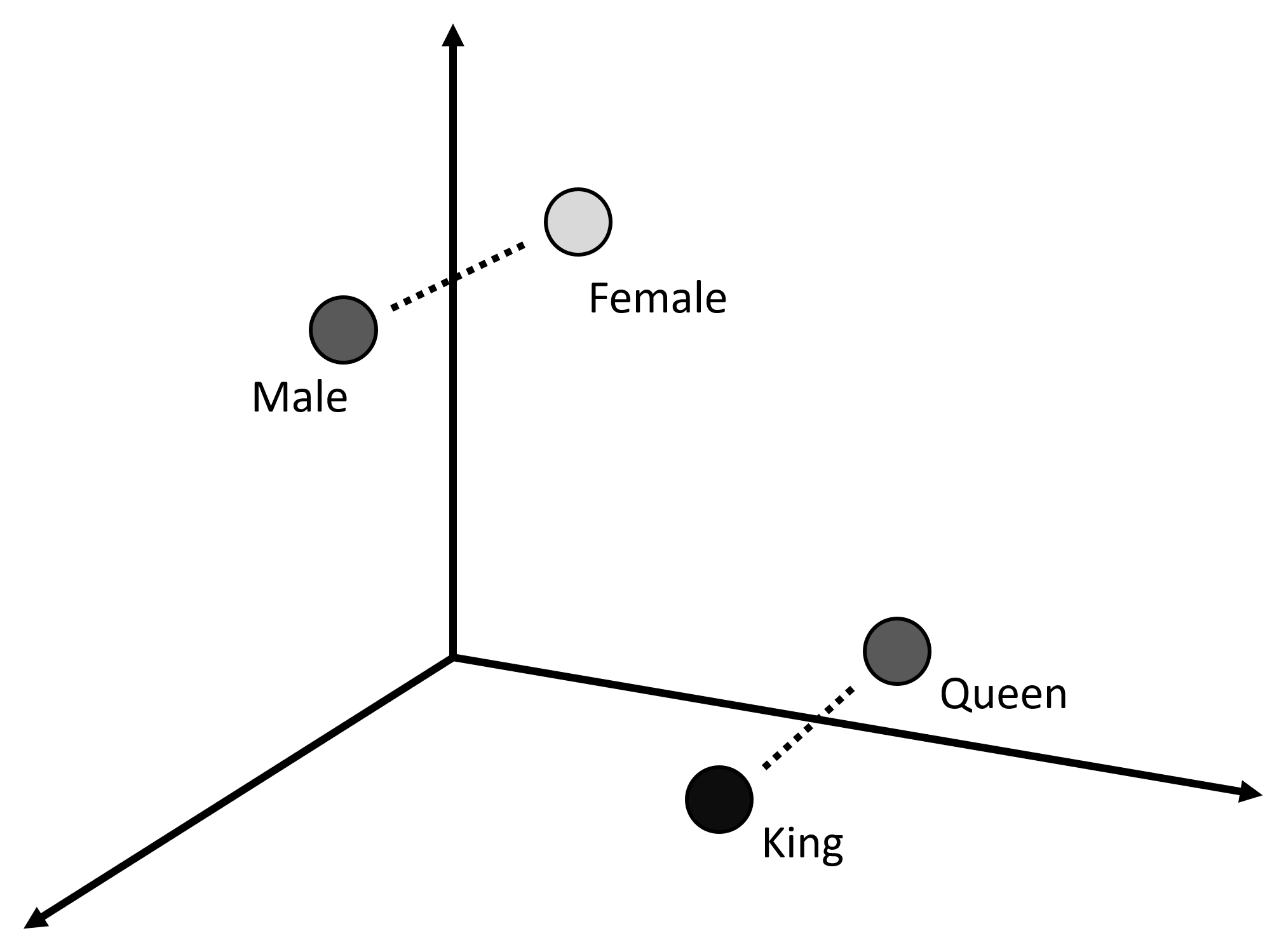}\hspace{0.3in}
}}
\caption{Embedded words in a continuous vector space. Words are represented as vectors with semantic meaning.}
\label{fig:word_embedding}
\end{figure}

To use categorical (or nominal) values like words in natural language processing in neural networks, the values should be projected into a continuous vector space, called {\em word embedding}, which captures relations among nominal values and represents them in a vector space \cite{Guo2016} as shown in Fig. \ref{fig:word_embedding}. In our experiments, feature embedding will also be included in our network architecture, which leads to better performance.  

Given a categorical variable $x \in {1, 2, \dots, T}$, where $T$ is the number of possible values that $x$ can take. Let $f$ be a simple function from $x$ to $e$, where $e$ is a one-hot vector in a $T$ dimensional space, and only the $x$th element of $e$ is one and the others are zeros. Then the vector representation of $x$ is defined as $W * e$, where $W$ is the embedding matrix with the shape of $(T \times D)$, and $D$ is the embedding dimension in which the categorical variable will be represented. In practice, $W * e$ might be implemented in a different way that the $x$th column of $W$ is selected, which is more efficient rather than the matrix-vector multiplication.  

The weight matrix $W$ represents weights connecting one-hot layer to embedding layer. Notice that weights can be initialized with random values and be trained just as other parameters in neural networks. Once a categorical variable is projected into a continuous vector, then the vector can be concatenated with other continuous input feature values, and the combined data travels to upper neural network layers.  

\section{Intrusion Detection based on LSTM}
Network intrusions have patterns according to their types. Generally, those patterns do not appear in a single packet, but can be dispersed for multiple packets. However, most of the previous machine learning methods for NIDS failed to address such characteristic, and they were not able to capture patterns that appear in multiple packets. For example, multi-layer Perceptron (MLP) performs intrusion detection with only one packet ignoring temporal dependency. Actually, if you want to detect DoS attack with MLP, it would be very tough because DoS is an attack to bring down a server by sending many packets, each of which is not very different from normal packets. This issue might not be limited to DoS attack, but also for other attack types. For more accurate intrusion detection, therefore, it is necessary to deal with multiple packets rather than a single packet.

\begin{figure}[h]
\centerline{\hbox{ 
\includegraphics[width=3.0in]{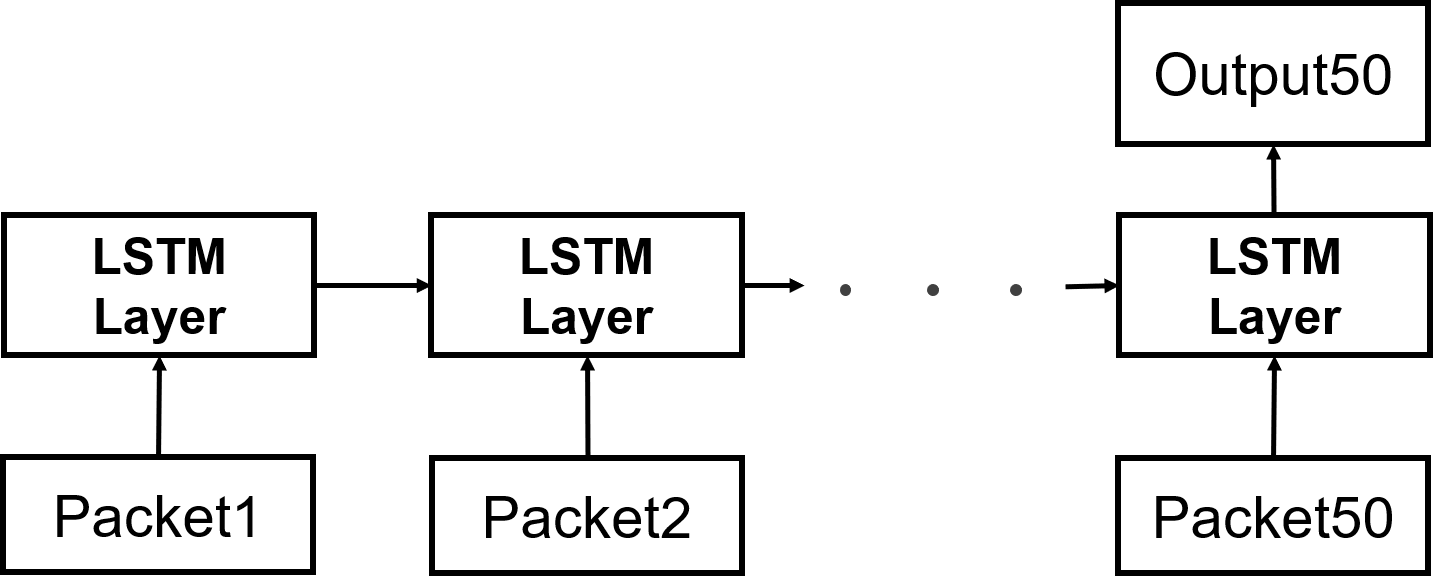}\hspace{0.3in}
}}
\caption{Intrusion detection through multiple packets}
\label{fig:detection_with_many_packets}
\end{figure}

In this study, we use LSTM to detect whether or not the current packet is normal considering the previous packets. The current packet and previous packets are put into LSTM as inputs as in Fig. \ref{fig:detection_with_many_packets}. 


However, there are several different ways to train the network or to construct our network architecture.
First, the network can be trained with the final label corresponding to the current packet or with all the labels to the current and the past packets. Second, the network can be constructed for binary classification (`normal' or `attack') or multiple classification (`normal' or several attack types). Even for binary classification, we can train a model for multiple classification and classify all types of attack as a single class `attack' as in binary classification. Further, the network can be constructed with or without embedding layer which is for categorical features.

\subsection{Many-To-Many Train vs. Many-To-One Train}
\begin{figure}[h]
\centerline{\hbox{ 
\includegraphics[width=2.0in]{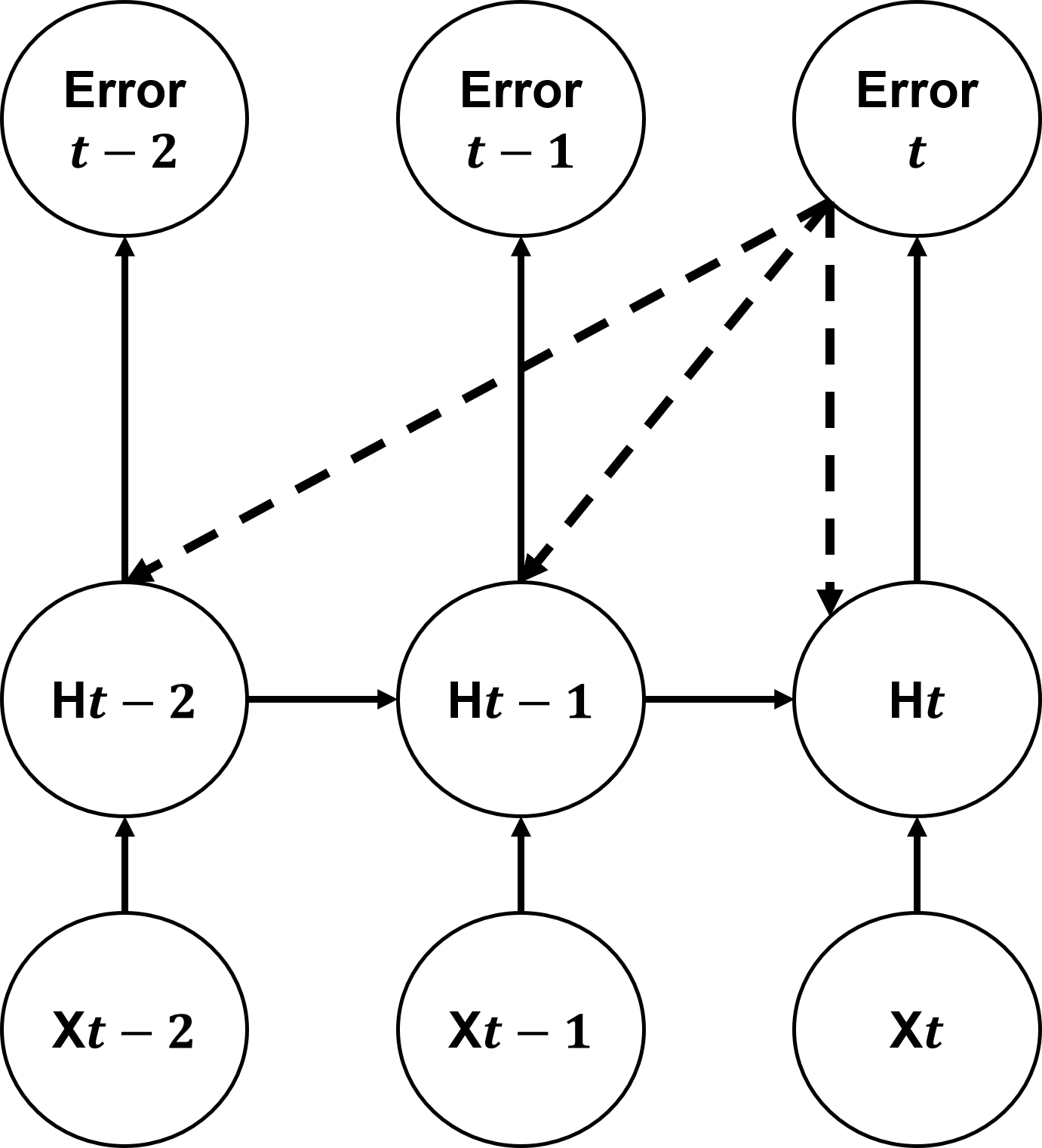}\hspace{0.5in}
\includegraphics[width=2.0in]{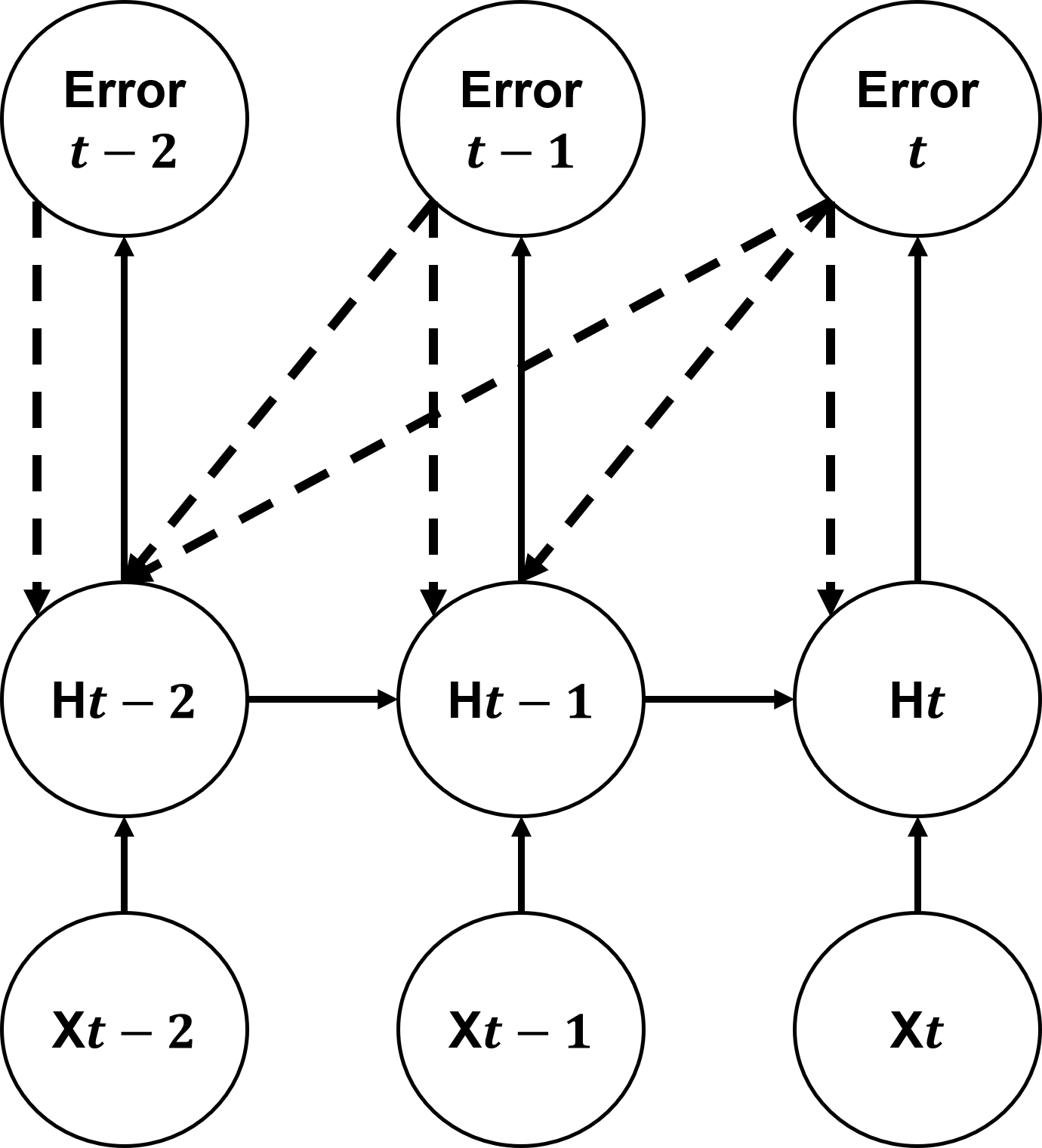}
}}
\vspace{0.1in}
\centerline{\hbox{(a) \hspace{2.3in} (b)}}
\vspace{0.1in}
\caption{Two learning methods: (a) M2O training learns only the last output, and (b) M2M training learns all the outputs in the sequence.}
\label{fig:many_one_train}
\end{figure}

Given sequential packets, NIDS is to perform many-to-one classification where the current input is classified using sequential packets as in Fig. \ref{fig:detection_with_many_packets}. That is, given many input steps, the output is determined only for the last step. Basically, RNNs like LSTM take one input packet at a time and yield a prediction output at each time step. Therefore, it is natural to train the model only with the last error as in many-to-one classification as in Fig. \ref{fig:many_one_train}(a), which is called many-to-one (M2O) training. However, it is possible to use all the errors for training as in Fig. \ref{fig:many_one_train}(b) if the labels are available, because the labels for the previous packets have some information which accelerate the training process, which is called many-to-many (M2M) training. 
That is, in the M2M strategy, not only the attack type of the target packet but also the attack type of the previous packets are learned together.
We compare two training approaches in experiments. 

\subsection{Multi Classes to Binary Class Detection}
\begin{figure}[h]
\centerline{\hbox{ 
\includegraphics[width=2.0in]{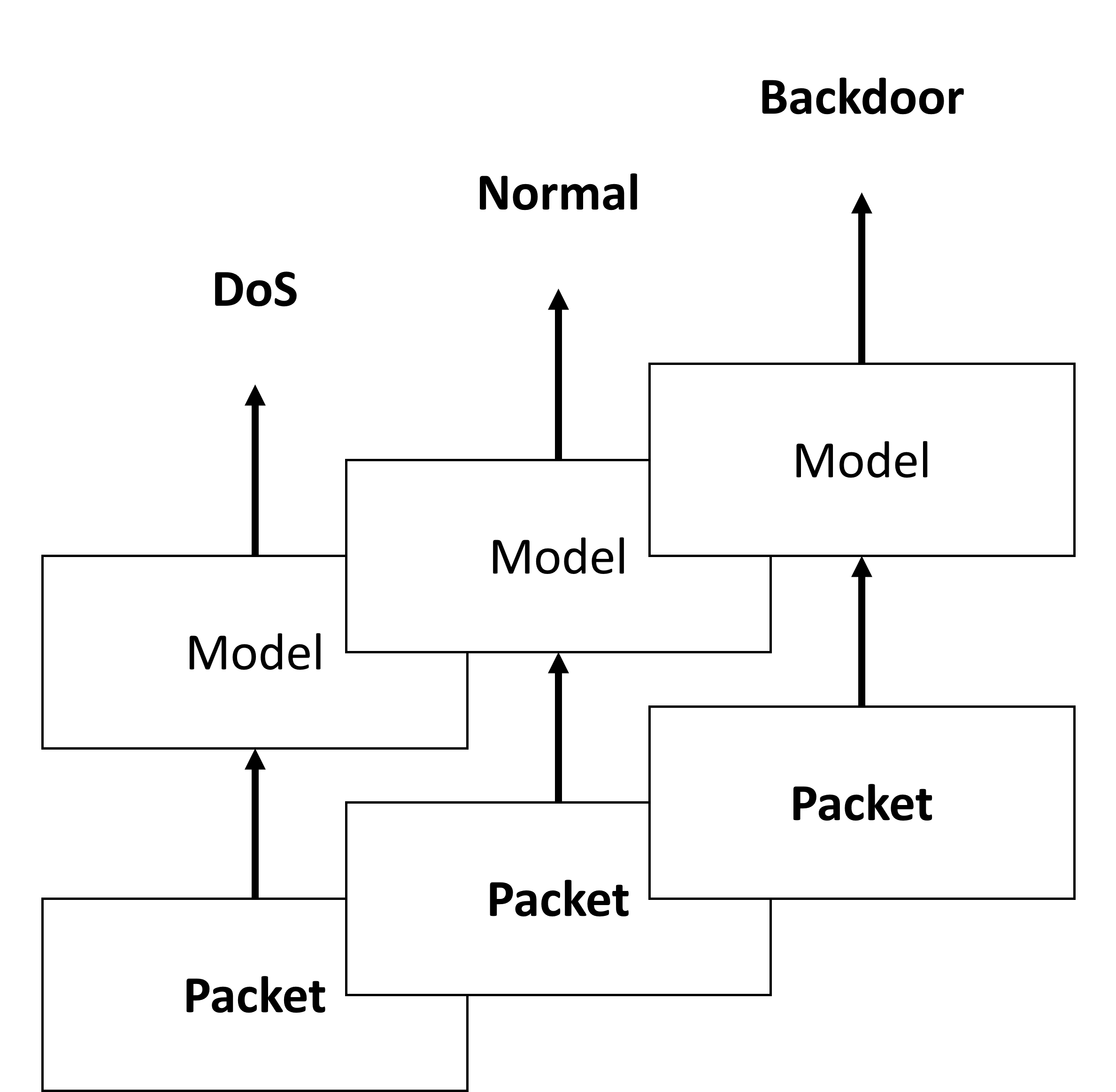}\hspace{0.5in}
\includegraphics[width=2.0in]{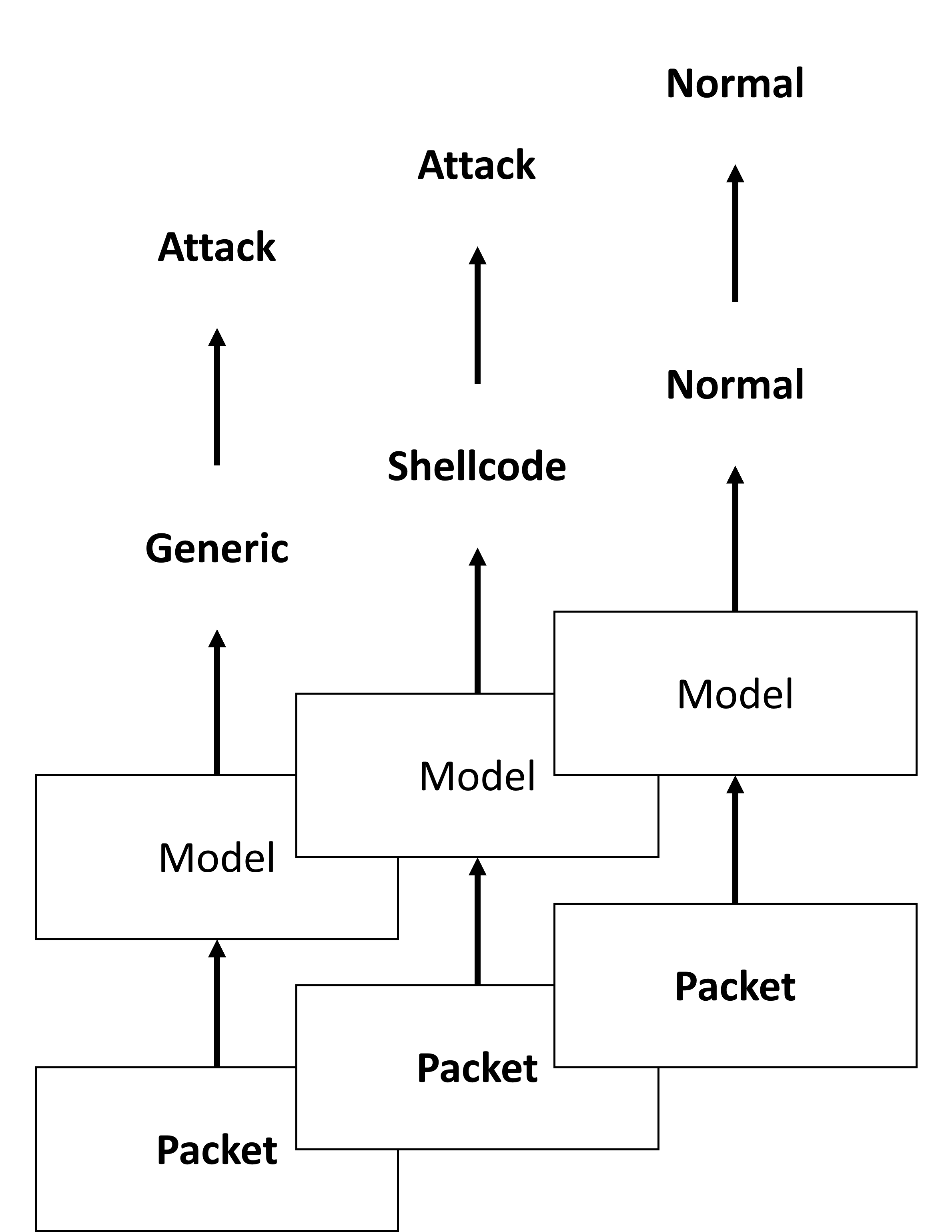}
}}
\vspace{0.1in}
\centerline{\hbox{(a) \hspace{2.3in} (b)}}
\vspace{0.1in}
\caption{M2B classification: (a) The model is trained to perform multi-classification,  (b) The prediction results are merged into binary classification results.}
\label{fig:m2b}
\end{figure}

There are various types of network intrusions, and multi-classification might be of interest. Sometimes, however, it would be interesting to classify a packet as normal or abnormal. For that, there are two approaches. First, various attack types of packets can be converted into `attack' before training. Then, the model is trained to classify a packet into binary results (`normal' or `attack'). Alternatively, without converting multiple attack types into a single class `attack', the model can be trained for multiple classification and the prediction results can be merged into binary classification results as in Fig. \ref{fig:m2b}. That is, basically the model performs multi-classification, but if the prediction is one of attack types, then it is classified to `attack'. We call it multi-to-binary (M2B) classification. 

\subsection{Detection With Feature Embedding}
Network packets (or connection) contain several nominal (or categorical) features like protocol and state. 
Each nominal feature indicates what role the packet plays and in what state it is. Each feature has characteristics that distinguish one packet from other packets with different nominal values. However, different values of one feature can have a very similar behavior by its functions. 
Therefore, simply replacing the nominal features with one-hot encoding vector may not be enough to represent packets. We apply the feature embedding technique to make each nominal feature a suitable vector in a continuous vector space according to the attack types.

As in NMT \cite{hchoi2017csl}, all nominal features are initialized to random vectors. With training, the vectors converge to appropriate points depending on packet’s attack types. For example, TCP and UDP often appear in the same attack type, so they are located closely in the vector space after training. With all embedded category features, our model could improve the detection performance utilizing relationships between nominal features.

\section{Experiments}
\subsection{Dataset}
To evaluate our proposed method for network intrusion detection system, we adopted the UNSW-NB15 dataset \cite{Moustafa2015}. UNSW-NB15 is an open dataset published by UNSW, a university in Australia, for network intrusion detection research in 2015. The KDD Cup 99 dataset used to be extensively used for network intrusion detection research in the past, but more recently UNSW-NB15 has been used because KDD Cup 99 does not contain much of the recent network hacking patterns \cite{Moustafa2015}. UNSW-NB15 consists of nine attack types and normal type as described in Table \ref{tbl:unsw_dist}. The dataset consists of 3 nominal, 2 binary, and 37 numerical features. The dataset splits into two sets: training(175,341 packets) and testing(82,332 packets). From the training set, 10\% of randomly selected samples are put aside and used for validation. 


In addition, the records of UNSW-NB15 are sorted in chronological order, which provides sequential patterns \cite{Moustafa2015}.

\begin{table}[h!] \centering	
\caption{UNSW-NB15 Dataset Attack Type}
    \label{tbl:unsw_dist}
    \begin{tabular}{| c | r | r |}
    \hline
    \textbf{ Category } & \textbf{ Train } & \textbf{ Test } \\ \hline \hline
    Total Records & 175,341 (100\%) & 82,332(100\%) \\ 
    \hline
   Normal & ~56,000(31.94\%)  & ~37,000(44.94\%) \\
   Analysis  & 2,000(1.14\%)  & 677(0.82\%) \\
   Backdoor  & 1,746(1.00\%)  &  583(0.71\%) \\
   Dos  & 12,264(6.99\%)  & 4,089(4.97\%) \\
   Exploits  & 33,393(19.04\%)  &  11,132(13.52\%) \\
   Fuzzers  & 18,184(10.37\%)  & 6,062(7.36\%) \\
   Generic  & 40,000(22.81\%)  &  18,871(22.92\%) \\
   ~Reconnaissance~  & 10,491(5.98\%)  & 3,496(4.25\%) \\
   Shellcode  & 1,133(0.65\%)  & 378(0.46\%) \\
   Worms  & 130(0.07\%)  & 44(0.05\%) \\
   \hline
   \end{tabular}
\end{table}

\subsection{Model Architecture}
\begin{figure}[h]
\centerline{\hbox{ 
\includegraphics[width=3.1in]{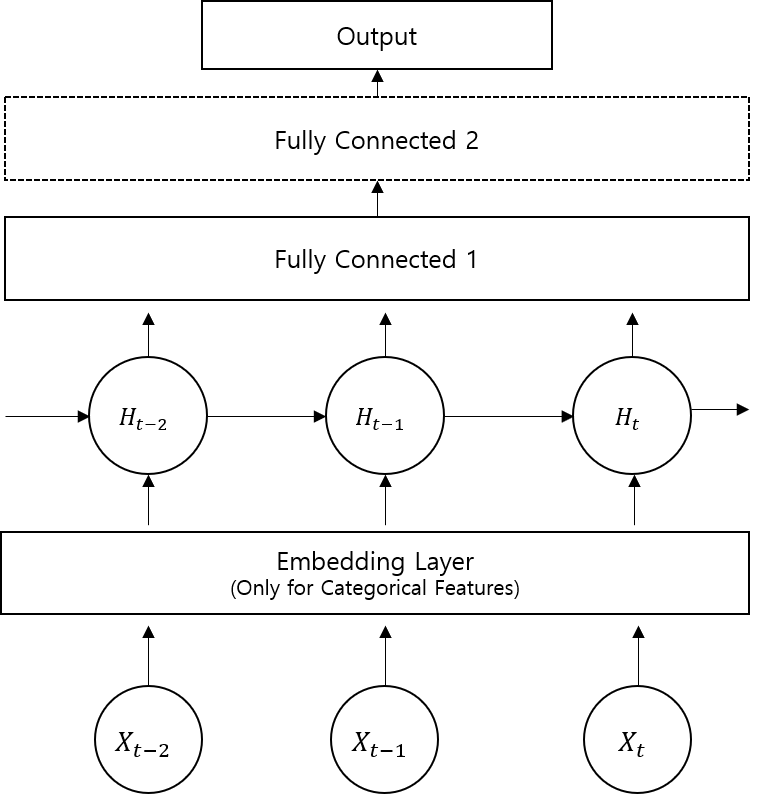}\hspace{0.3in}
}}
\caption{Model Architecture: embedding, LSTM, and fully connected layers. `Fully Connected 2' is used only for binary classification.}
\label{fig:LSTM_model}
\end{figure}
Our model is composed of 3 types of layers: embedding, LSTM, and fully connected layers. The embedding layer is only for nominal features of an input, and continuous features are set aside. 3 nominal features (proto, service, and state) are mapped to 5, 3, and 2 dimensional vectors, respectively. These output vectors are concatenated to continuous features and travel to the next layer in the model. The LSTM layer is composed of hidden state with 100 nodes. The fully connected layer is of size 50 with dropout. As activation function, leaky ReLU \cite{He2015} is applied for non-linear transformation. In case of binary classification, the second fully connected layer is added with size of 10 nodes. In Fig. \ref{fig:LSTM_model}, the dotted line indicates the layer working only in case of binary classification.

\subsection{Evaluation Metrics}
As evaluation metrics, we used accuracy (AC) and F1-score (F1). Given true positive (TP), true negative (TN), false positive (FP), and false negative (FN), AC and F1  are respectively calculated by 
\begin{eqnarray}
AC &=& \frac{TP + TN}{ TP+TN+FP+FN}, \\
F1 &=& \frac{2P*R}{P + R},
\end{eqnarray}
where P and R stand for precision and recall, respectively as follows. 
\begin{eqnarray}
P &=& \frac{TP}{TP + FP}, \\
R &=& \frac{TP}{TP + FN}. 
\end{eqnarray}
As the harmonic mean of precision and recall, F1-score provides a better evaluation measure than accuracy especially for imbalanced data.

\subsection{Experiment Results}
We evaluate many combination of training configurations on LSTM with feature embedding. First, the LSTM model is trained in two ways as we described above. One is learning from errors of every output (M2M) and the other one is learning only from the error of the last output (M2O). In addition, for binary-classification, we add `multi-classification to binary-classification' (M2B) which trains a multi-classification model and converts all of malicious labels and outputs of model to the same label `attack'. Finally, feature embedding (EMB) is applied to every models.

\begin{table}[h!] \centering	
\caption{Binary-classification LSTM Model results for test data. Validation results are in the parenthesis.}
\label{tbl:LSTM_comp_binary}
\begin{tabular}{| l| c | c | c |}
\hline
\textbf{Model} &  \textbf{Sequence Length}  & \textbf{ Accuracy } & \textbf{ F1 Score } \\ 
\hline
 ANN \cite{Suleiman2018} & - & 81.91 & 95.2 \\
 RepTree \cite{Belouch2017} & - & 88.95 & - \\
 Random Forest \cite{Vinayakumar2019} & - & 90.3 & 92.4 \\
 MLP & - & 83.55 (94.00) & 86.89 \\
 LSTM(M2M) &  110 & 98.68 (99.88) & 99.16 \\ 
 LSTM(M2O) &  310 & 98.49 (97.99) & 98.90 \\  
 LSTM(M2M M2B)  &  130 & 98.29 (99.84) & 98.43 \\ 
 LSTM(M2O M2B)  &  210 & 99.42 (98.07) & 99.47 \\ 
 LSTM(M2M + EMB) & 270 & ~\textbf{99.72 (99.97)}~ & \textbf{99.75} \\ 
 LSTM(M2O + EMB) &  90 & 99.52 (97.82) & 99.56 \\ 
 LSTM(M2M M2B + EMB) &  110 & 99.53 (99.93) & 99.67 \\ 
 LSTM(M2O M2B + EMB) &  110 & 98.83 (98.02) & 98.93 \\  
 \hline
\end{tabular}
\end{table}

\begin{table}[h!] \centering	
\caption{Multi-classification LSTM Model results. Validation results are in the parenthesis.}
    \label{tbl:LSTM_comp_multi}
    \begin{tabular}{| l| c | c |}
    \hline
    \textbf{Model} &  \textbf{Sequence Length}  & \textbf{ Accuracy } \\ 
    \hline
 Random Forest \cite{Vinayakumar2019} & - & 75.5\\
 RepTree \cite{Belouch2017} & - & 81.28 \\
 MLP & - & 72.81 (79.32) \\
 LSTM(M2M) &   20 & 84.78 (85.52)\\ 
 LSTM(M2O) &  250 & 83.45 (82.72)\\ 
 LSTM(M2M + EMB) &   30 &~\textbf{86.98 (88.50)}~\\ 
 LSTM(M2O + EMB) &  150 & 85.93 (83.00)\\ 
 \hline
   \end{tabular}
\end{table}

MLP model and LSTM models have apparent differences in terms of performances as summarized in Tables \ref{tbl:LSTM_comp_binary} and \ref{tbl:LSTM_comp_multi}. The MLP model shows the accuracy of 83.55\% and 72.81\% for binary-classification and multi-classification, respectively. The corresponding F1 score for the binary case is 86.89\%. The LSTM models show the accuracy over 98\% in binary-classification (F1 score of 99.75\%) and 83\% in multi-classification. The LSTM models outperform because LSTM can capture the temporal dependency presented in sequence of packets, while MLP cannot. In addition, our LSTM models outperform the previous works \cite{Suleiman2018,Belouch2017,Vinayakumar2019}

\begin{figure}[h]
\centerline{\hbox{ 
\includegraphics[width=3.4in]{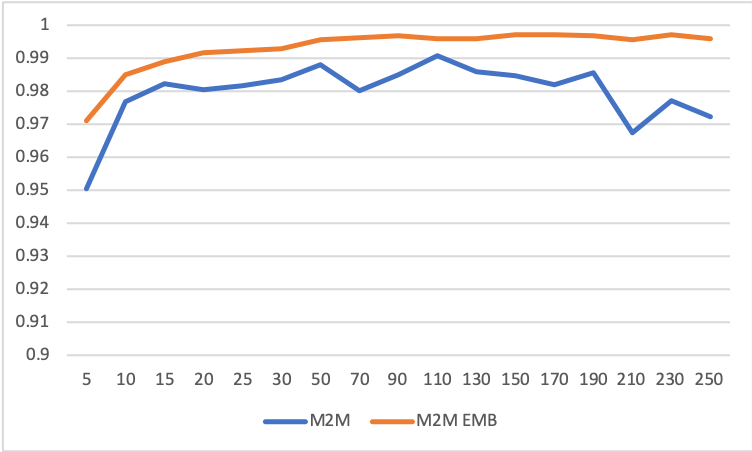}\hspace{0.05in}}}
\caption{Binary-classification accuracy graphs on the validation data: M2M, and M2M with embedding. The horizontal axis indicates the length of sequence.}
\label{fig:binary_graph}
\end{figure}

\begin{figure}[h]
\centerline{\hbox{ 
\includegraphics[width=3.4in]{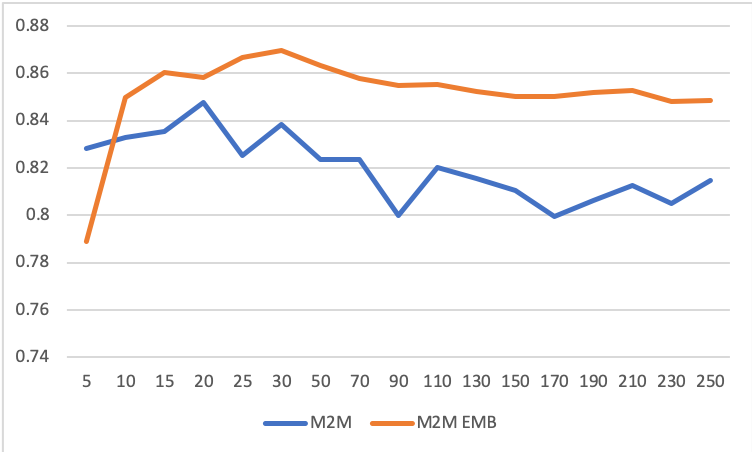}}}
\caption{Multi-classification accuracy graphs on the validation data: M2M, and M2M with embedding. The horizontal axis indicates the length of sequence.}
\label{fig:multi_graph}
\end{figure}

Among the LSTM models, the M2M+EMB model achieved the highest performance for both binary and multiple classification tasks. It is because categorical features includes distinguishable information and feature embedding is efficient to capture the information for neural networks. 
Actually, when comparing EMB models to corresponding non-EMB models, the EMB models have better performance (around 1\% higher for binary classification and 2\% higher for multi-classification) and more stable results as shown in Figs. \ref{fig:binary_graph} and \ref{fig:multi_graph}.

In addition, for binary-classification, M2B can be applied, but makes no significant influence on performance. The results of M2B and non-M2B models are almost the same. 

\begin{figure}[h]
\centerline{\hbox{ 
\includegraphics[width=3.75in]{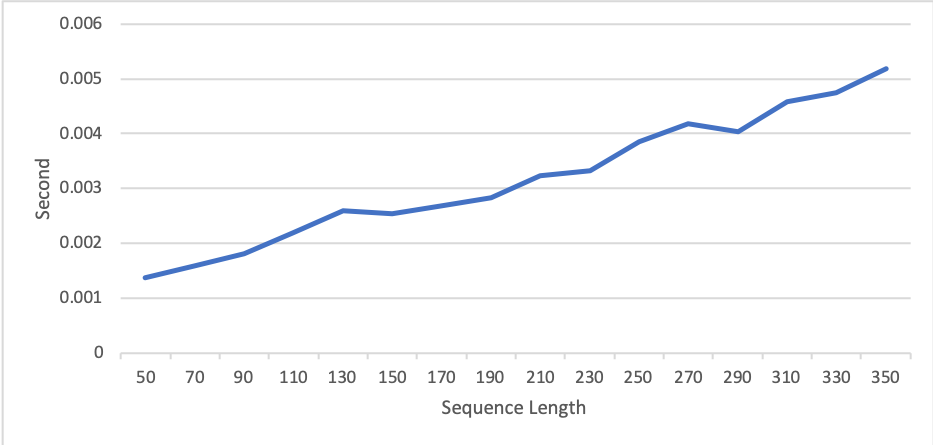}}}
\caption{Prediction time in seconds per sequence with various sequence lengths.}
\label{fig:time_consumption}
\end{figure}

For practical consideration, we checked the prediction time with different sequence length in the model, and the results are summarized in Fig. \ref{fig:time_consumption}, where we can see that the prediction time is linear to the sequence length.

\section{Conclusion}
In this paper, we proposed and experimented several IDS models based on LSTM and feature embedding. Evaluation was based on the UNSW-NB15 dataset which is suitable to reflect latest network traffic patterns. LSTM outperformed MLP with a significant margin (around 16\% point or 13\%) in accuracy and F1 score. Among LSTM models, the one with feature embedding was the best, since the embedding technique could capture categorical information which is crucial for attack recognition. 

We expect that real-time detection is possible in practice. Our future work includes making the model compatible with embedded system and Internet of things (IoT) by reducing the model complexity and shortening the necessary sequence length.

\section*{Acknowledgement}
This research was supported by Basic Science Research Program through the National Research Foundation of Korea(NRF) funded by the Ministry of Education (2017R1D1A1B03033341), and by Institute for Information \& communications Technology Promotion(IITP) grant funded by the Korea government(MSIT) (No. 2018-0-00749, Development of virtual network management technology based on artificial intelligence).

%
%
\bibliographystyle{splncs04}

\end{document}